# BrainNext: A General-Purpose Self-Supervised Foundation Model for Brain MRI Analysis

Moona Mazher[1], Abdul Qayyum[2], Steven A. Niederer[2,3], Daniel C. Alexander[1]

[1]UCL Hawkes Institute, Department of Computer Science, University College London, London, United Kingdom

[2]National Heart & Lung Institute, Imperial College London, London, United Kingdom

[3]Division of Cardiovascular Medicine, Stanford Cardiovascular Institute, Stanford University, California, USA

## Abstract

Foundation models pretrained using self-supervised learning have transformed computer vision by learning transferable representations from large-scale unlabeled data. However, existing foundation models for neuroimaging remain limited by task-specific training, slice-based learning strategies, or relatively small pretraining datasets, restricting their generalizability across diverse brain MRI applications. In this work, we present BrainNext, a general-purpose self-supervised foundation model for volumetric brain MRI analysis. BrainNext combines masked autoencoder (MAE) pretraining with a native three-dimensional Bi-Directional xLSTM-UNet architecture to learn rich anatomical representations from 60,551 unlabeled brain MRI examinations spanning multiple MRI modalities. The pretrained model is subsequently adapted to downstream tasks through lightweight task-specific fine-tuning. We evaluate BrainNext on the Foundation Models for Medical Imaging (FOMO) 2025 Method Track, encompassing classification, segmentation, and brain-age estimation, where it achieved second place overall and ranked first in the meningioma segmentation task on the official FOMO 2025 challenge leaderboard, demonstrating strong transferability across heterogeneous neuroimaging tasks. These results highlight the potential of large-scale self-supervised pretraining to learn robust and transferable volumetric representations, establishing BrainNext as a scalable foundation model for diverse brain MRI applications.

## 1. Introduction

Magnetic resonance imaging (MRI) has become an indispensable tool [1,2] for the diagnosis, prognosis, and longitudinal monitoring of neurological disorders, including stroke, brain tumours, Alzheimer's disease, multiple sclerosis, and other neurodegenerative conditions. The widespread adoption of MRI in both clinical practice and large-scale research studies has generated vast collections of neuroimaging data, creating unprecedented opportunities for artificial intelligence (AI) to advance automated brain image analysis. Deep learning methods have achieved remarkable success across diverse neuroimaging applications [3], including disease classification, anatomical and pathological segmentation, quantitative biomarker extraction, and brain-age estimation. Despite these advances, most existing models remain highly task-specific, requiring large annotated datasets and independent training for each downstream application. Such dependence on manual annotation limits scalability and restricts the development of generalizable models that can be readily transferred across different clinical tasks.

Self-supervised learning (SSL) has recently emerged as an effective paradigm for learning transferable representations directly from unlabeled imaging data [4-8]. Rather than relying on manually curated labels, SSL exploits intrinsic image structure through pretext objectives such as contrastive learning, self-distillation, predictive representation learning, and masked image modelling. These approaches have demonstrated impressive transferability across downstream tasks while substantially reducing annotation requirements. More recently, this paradigm has evolved into foundation models, where a single pretrained model learns general-purpose representations that can be efficiently adapted to multiple downstream applications through lightweight task-specific fine-tuning. Foundation models have rapidly transformed natural language processing and computer vision and are increasingly being explored for medical imaging. [9-11, 27]

Recent studies have extended foundation model paradigms to neuroimaging. Brain-specific foundation models have employed diverse self-supervised objectives, including self-distillation, contrastive learning, and masked image modelling, together with different backbone architectures ranging from two-dimensional vision transformers to native three-dimensional convolutional networks [14-19]. While these studies have demonstrated encouraging transfer performance across downstream applications, several challenges remain. Many existing approaches process volumetric MRI as independent two-dimensional slices, limiting their ability to explicitly model three-dimensional anatomical context. Furthermore, existing foundation models are frequently evaluated using different pretraining datasets, downstream tasks, and experimental protocols, making direct comparison difficult. Furthermore, relatively few studies have been evaluated under a common benchmark using identical pretraining data and downstream tasks. Standardized evaluation is essential for understanding how different self-supervised learning strategies translate into generalizable representations across diverse neuroimaging applications. The FOMO benchmark addresses this limitation by providing a unified evaluation protocol for brain MRI foundation models. These limitations highlight the need for scalable three-dimensional foundation models that learn transferable volumetric representations and are systematically evaluated across diverse neuroimaging tasks within a unified benchmark.

To address these challenges, we present BrainNext, a general-purpose self-supervised foundation model for volumetric brain MRI analysis. BrainNext combines masked autoencoder (MAE) pretraining with a native three-dimensional Bi-Directional xLSTM-UNet architecture to learn transferable anatomical representations from 60,551 unlabeled brain MRI examinations. Unlike conventional supervised models that are optimized for a single application, BrainNext is designed as a reusable representation learner whose pretrained encoder can be efficiently adapted to multiple downstream tasks through lightweight fine-tuning. We evaluate BrainNext using the standardized FOMO 2025 Method Track, which provides a unified benchmark for assessing foundation models under limited-label and out-of-domain conditions across the three principal categories of medical image analysis: classification, segmentation, and regression.

The main contributions of this work are summarized as follows:

- We present BrainNext, a general-purpose self-supervised foundation model for volumetric brain MRI based on masked autoencoder pretraining and a native three-dimensional Bi-Directional xLSTM-UNet architecture.
- We perform large-scale self-supervised pretraining on 60,551 unlabeled brain MRI examinations, enabling transferable volumetric representations without manual annotation.
- We comprehensively evaluate BrainNext on the standardized FOMO 2025 benchmark across the three principal categories of medical image analysis: infarct classification, meningioma segmentation, and brain-age estimation.
- We will publicly release the pretrained model weights and inference code to facilitate reproducible research and future development of foundation models for brain MRI analysis.

## 2. Related Work

### 2.1 Self-supervised learning for medical imaging

Deep learning methods for brain MRI have traditionally been developed for individual tasks such as classification, segmentation, or regression [3-5]. Although task-specific supervised models can achieve strong performance, their development depends on labelled datasets that are costly to curate and often limited in size. Performance may also degrade when models are evaluated across different institutions, scanners, acquisition protocols, or patient populations. Self-supervised learning addresses these limitations by learning representations from unlabeled images before adaptation to a downstream task.

Several self-supervised paradigms have been explored in medical imaging, including contrastive learning, self-distillation, masked image modelling, and predictive representation learning. Contrastive approaches such as SimCLR and MoCo learn representations by encouraging agreement between augmented views of the same image, whereas self-distillation methods such as DINO learn representations through teacher–student consistency [6-11]. Masked autoencoders instead reconstruct missing image regions from the visible context, providing a natural objective for volumetric medical images in which spatial structure is central. The original MAE framework demonstrated that high masking ratios can support efficient representation learning, and subsequent work has adapted masked reconstruction to three-dimensional medical imaging.

Large-scale studies have further examined how pretraining data, architecture, and objective design influence downstream transfer. AMAES [17] used masked autoencoder pretraining on 44,756 public brain MRI volumes and demonstrated improvements across several three-dimensional segmentation settings. Related work revisiting MAE pretraining within a residual encoder U-Net [13] framework showed that large-scale pretraining on approximately 39,000 brain MRI volumes could improve performance across multiple segmentation datasets. OpenMind [18] subsequently introduced a large public resource containing approximately 114,000 unlabeled three-dimensional brain MRI volumes to support standardized comparisons of self-supervised methods. Collectively, these studies established masked reconstruction as a practical approach for learning volumetric brain MRI representations, although much of the evaluation remained focused on segmentation.

### 2.2 Foundation models for brain MRI

The emergence of large and heterogeneous neuroimaging datasets has enabled the development of foundation models designed to support multiple downstream applications from a shared pretrained representation. These models differ in their pretraining objectives, input dimensionality, architectures, and adaptation strategies.

BrainFound [14] adapted DINOv2 self-distillation to brain MRI and processed volumetric examinations as sequences of two-dimensional axial slices. The model demonstrated transfer across disease classification and anatomical and pathological segmentation tasks. Its slice-based design enabled the use of a large pretrained vision-transformer backbone, but it did not explicitly model the full three-dimensional volume within a native volumetric encoder.

BrainIAC [15] used self-supervised contrastive pretraining on 48,965 multiparametric brain MRI scans and was evaluated across a broad range of downstream applications using limited task-specific supervision. Its results demonstrated the value of brain-specific pretraining relative to conventional supervised learning and transfer from more general biomedical models.

Other studies have investigated native three-dimensional MRI foundation models. Triad [16] pretrained an autoencoder-based model on 131,170 three-dimensional MRI volumes from multiple anatomical regions and evaluated transfer across segmentation, classification, and registration. AMAES [17] and related MAE-based approaches focused more specifically on three-dimensional brain MRI segmentation, while BrainAgeNeXt [19] investigated pretrained representations for brain-age modelling. Together, these studies demonstrate increasing interest in volumetric MRI pretraining, but they also show that models are often optimized or evaluated primarily for a restricted family of downstream tasks.

### 2.3 Benchmarking brain MRI foundation models

Direct comparison of brain MRI foundation models is difficult because studies frequently use different pretraining datasets, downstream cohorts, architectures, label fractions, and evaluation protocols. The FOMO25 challenge was designed to provide a standardized and clinically relevant benchmark under limited-label and out-of-domain conditions. Participants pretrained a common foundation model and adapted it independently to three representative task categories: infarct classification, meningioma segmentation, and brain-age regression. The Method Track restricted pretraining to FOMO60K, comprising 60,529 MRI scans from 13,900 sessions and 11,187 subjects across 16 public sources, thereby reducing differences arising from access to external data [28].

The challenge findings showed that self-supervised pretraining generally improved out-of-domain performance relative to supervised baselines trained on the same limited labelled datasets. However, the results also indicated that no single pretraining objective performed best across all downstream tasks. Local reconstruction objectives, such as MAE, tended to favour segmentation, whereas hybrid reconstruction-contrastive objectives tended to perform better for classification. Model size and longer pretraining duration were not consistently associated with improved overall performance. These findings highlight the importance of evaluating foundation models across multiple task families rather than drawing conclusions from a single downstream benchmark.

### 2.4 Positioning of BrainNext

BrainNext was developed within the FOMO25 Method Track as a general-purpose self-supervised foundation model for native volumetric brain MRI. It combines masked autoencoder pretraining with a three-dimensional Bi-Directional xLSTM-UNet backbone. The model contains approximately 45 million parameters and was pretrained on FOMO60K using $96 \times 96 \times 96$ voxel patches, with each MRI sequence treated independently during pretraining. The same pretrained encoder was then adapted separately for infarct classification, meningioma segmentation, and brain-age regression using task-specific heads and fine-tuning strategies.

BrainNext differs from the authors' previously published BrainFound [15] model in both dimensionality and pretraining strategy. BrainFound uses DINOv2-based self-distillation with a two-dimensional slice-based representation, whereas BrainNext uses masked reconstruction within a native three-dimensional architecture. BrainNext is therefore not presented as a direct three-dimensional implementation of BrainFound, but as a separate foundation-model framework designed to learn volumetric representations and transfer across classification, segmentation, and regression. Its evaluation on the FOMO25 benchmark provides a controlled assessment under few-shot adaptation and substantial domain shift, where the Dolphins_creators submission ranked second overall in the Method Track and first on the meningioma segmentation task. Table 1 summarizes representative brain MRI foundation models and highlights the differences in self-supervised learning strategy, backbone architecture, and downstream applications.

Table 1. Comparison of representative self-supervised foundation models for brain MRI. Models are compared according to their self-supervised learning objective, backbone architecture, native 3D capability, and primary downstream tasks reported in the original publications.

| **Model** | **SSL Objective** | **Backbone** | **Native 3D** | **Primary Downstream Tasks** |
|---|---|---|---|---|
| BrainFound | DINOv2 | ViT | ✗ | Classification, Segmentation |
| BrainIAC | Contrastive | CNN | ✓ | Classification, Regression |
| Triad | Autoencoder | CNN | ✓ | Classification, Segmentation, Registration |
| AMAES | MAE | U-Net | ✓ | Segmentation |
| BrainAgeNeXt | MAE | CNN | ✓ | Brain-age Estimation |
| **BrainNext (Ours)** | MAE | Bi-Directional xLSTM-UNet | ✓ | Classification, Segmentation, Regression |

## 3. BrainNext

### 3.1 Overview

BrainNext is a general-purpose self-supervised foundation model designed for volumetric brain MRI analysis. The proposed framework follows a two-stage learning paradigm consisting of large-scale self-supervised pretraining followed by task-specific fine-tuning, as illustrated in Figure 1. During the first stage, the model learns transferable anatomical representations from 60,551 unlabeled brain MRI examinations using a masked autoencoder (MAE) [12] objective. Unlike supervised learning, this stage does not require manual annotations and instead exploits the intrinsic spatial structure of brain MRI volumes to learn robust feature representations. In the second stage, the pretrained encoder is adapted to downstream applications through lightweight task-specific heads. The effectiveness of the learned representations is evaluated on the FOMO 2025 [29] Method Track across three representative categories of medical image analysis: infarct classification, meningioma segmentation, and brain-age estimation.

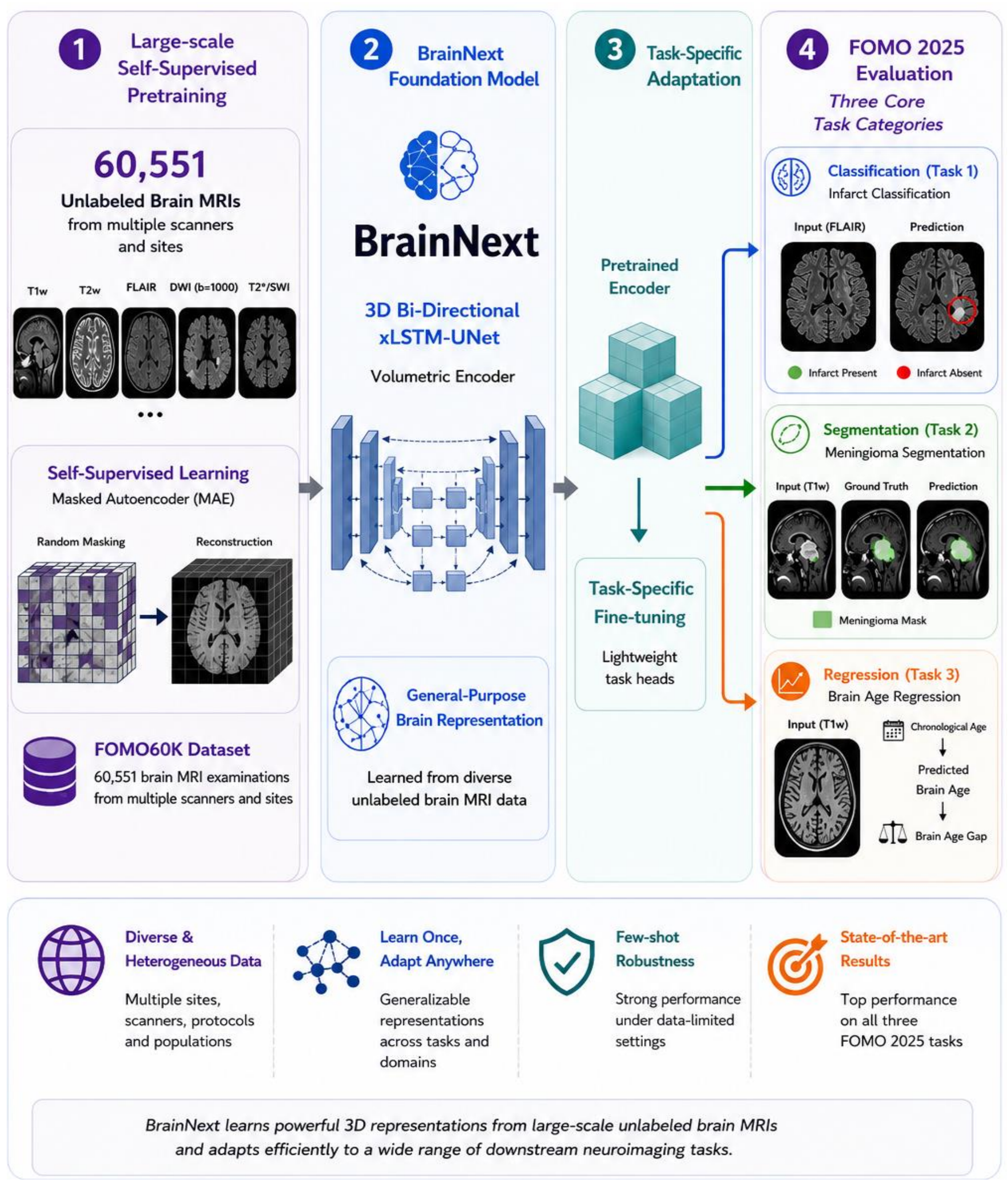

Figure 1. Overview of the BrainNext framework. (1) BrainNext is pretrained in a self-supervised manner on 60,551 unlabeled brain MRI examinations from the FOMO60K dataset using a masked autoencoder (MAE) objective. (2) The resulting BrainNext foundation model employs a native three-dimensional Bi-Directional xLSTM-UNet architecture to learn transferable volumetric representations from diverse brain MRI data. (3) The pretrained encoder is subsequently adapted to downstream applications through lightweight task-specific fine-tuning using task-specific prediction heads. (4) BrainNext is evaluated on the FOMO 2025 Method Track across three representative neuroimaging tasks: infarct classification, meningioma segmentation, and brain-age estimation.

### 3.2 Dataset and Data Preprocessing

BrainNext is pretrained using the FOMO60K dataset, comprising 60,551 unlabeled brain MRI examinations collected from multiple public neuroimaging datasets. The diversity of imaging protocols, scanners, and subject populations enables the model to learn robust and transferable anatomical representations. Prior to training, all MRI volumes are resampled to a common isotropic voxel spacing of 1 mm and intensity-normalized using z-score normalization. To facilitate efficient three-dimensional learning, images are processed using patch-based sampling. During fine-tuning, extensive data augmentation, including random rotations, flipping, scaling, Gaussian blurring, and additive Gaussian noise, is applied to improve robustness to anatomical variability and imaging heterogeneity.

### 3.3 BrainNext Architecture

BrainNext is built upon a native three-dimensional Bi-Directional xLSTM-UNet architecture that combines the hierarchical feature extraction capability of U-Net [20] with the long-range dependency modelling of bidirectional xLSTM [23] modules. The network consists of five encoder stages and five decoder stages. Each encoder stage contains consecutive 3D convolutional layers followed by bidirectional xLSTM blocks. The convolutional layers extract fine-grained local anatomical features, whereas the bidirectional xLSTM modules capture long-range contextual dependencies throughout the volumetric feature maps. Residual connections are incorporated to facilitate feature propagation and improve optimization stability.

Unlike purely convolutional architectures, the Bi-Directional xLSTM modules capture long-range volumetric dependencies while preserving hierarchical local anatomical features extracted by the convolutional encoder. This combination enables efficient modelling of both local and global anatomical context within three-dimensional MRI volumes.

Given the encoder feature map at level $l$,

$$F_l,$$

the bidirectional xLSTM computes forward and backward contextual representations

$$\overrightarrow{H}_l = \text{xLSTM}_{forward}(F_l),$$
$$\overleftarrow{H}_l = \text{xLSTM}_{backward}(F_l).$$

The two representations are concatenated and fused through a learnable projection with a residual connection,

$$H_l = F_l + \phi([\overrightarrow{H}_l, \overleftarrow{H}_l]),$$

where $\phi(\cdot)$ denotes a learnable fusion layer. During decoding, hierarchical encoder representations are combined with progressively upsampled decoder features through skip connections,

$$F_l^{dec} = U(F_{l+1}^{dec}) \oplus H_l,$$

where $U(\cdot)$ denotes upsampling and $\oplus$ denotes feature concatenation. The complete BrainNext framework contains approximately 45 million trainable parameters and performs end-to-end volumetric representation learning directly on three-dimensional brain MRI data.

**3.4 Self-Supervised MAE Pretraining**

BrainNext adopts the Masked Autoencoder (MAE) framework for large-scale self-supervised representation learning. Given an input MRI volume

$$x \in \mathbb{R}^{H \times W \times D \times C},$$

where $H$, $W$, and $D$ denote the spatial dimensions and $C$ represents the number of MRI channels, the image is partitioned into a sequence of non-overlapping volumetric patches,

$$x = \{x_1, x_2, \dots, x_N\},$$

where $N$ denotes the total number of patches. A random subset of patches,

$$\mathcal{M} \subset \{1, \dots, N\},$$

is masked during training, while only the remaining visible patches,

$$x_v = x \setminus \mathcal{M},$$

are processed by the encoder,

$$z = E(x_v).$$

A lightweight decoder reconstructs the missing patches,

$$\hat{x} = D(z),$$

and the model is optimized by minimizing the reconstruction error over the masked patches,

$$\mathcal{L}_{MAE} = \frac{1}{|\mathcal{M}|}\sum_{i\in\mathcal{M}} \| x_i - \hat{x}_i \|_2^2.$$

This reconstruction objective enables BrainNext to learn transferable anatomical representations by exploiting both local image appearance and global structural relationships without requiring manual annotations. Multiple MRI sequences are treated as separate input channels, allowing the network to jointly learn complementary information across imaging modalities [11].

## 3.5 Downstream Task Adaptation

After completion of self-supervised pretraining, the MAE reconstruction decoder is discarded while the pretrained encoder is retained. The downstream model is initialized as

$$\Theta = \{\Theta_{\text{enc}}, \Theta_{\text{head}}\},$$

where $\Theta_{\text{enc}}$denotes the pretrained encoder parameters and $\Theta_{\text{head}}$represents the parameters of the task-specific prediction head initialized for each downstream application.

BrainNext is adapted to three representative neuroimaging tasks: infarct classification, meningioma segmentation, and brain-age estimation. For infarct classification, encoder features are aggregated using global average pooling followed by fully connected layers with dropout to generate binary predictions. For meningioma segmentation, the pretrained encoder is coupled with a five-level U-Net decoder to produce voxel-wise segmentation masks. For brain-age estimation, encoder features are aggregated using global average pooling and processed through fully connected regression layers to predict chronological brain age.

Each downstream task is optimized using its corresponding objective function,

$$\mathcal{L} = \begin{cases} \mathcal{L}_{\text{BCE}}, & \text{for infarct classification,} \\ \mathcal{L}_{\text{Dice}} + \mathcal{L}_{\text{CE}}, & \text{for meningioma segmentation,} \\ \mathcal{L}_{\text{MSE}}, & \text{for brain-age estimation,} \end{cases}$$

where $\mathcal{L}_{\mathrm{BCE}}$denotes the binary cross-entropy loss, $\mathcal{L}_{\mathrm{Dice}}$is the Dice loss, $\mathcal{L}_{\mathrm{CE}}$is the cross-entropy loss, and $\mathcal{L}_{\mathrm{MSE}}$is the mean squared error loss.

This unified adaptation strategy enables a single pretrained encoder to support multiple categories of medical image analysis while preserving the learned volumetric representations.

### 3.6 Training and Implementation

Self-supervised pretraining is performed for 1000 epochs using the AdamW [24] optimizer with an initial learning rate of $2 \times 10^{-4}$and a cosine annealing learning-rate schedule. Fine-tuning is conducted independently for each downstream application using five-fold cross-validation. Infarct classification and brain-age estimation are trained for 500 epochs, whereas meningioma segmentation is optimized for 1000 epochs. Model selection is based on the best validation performance for each task using validation loss, Dice score, or mean absolute error respectively.

BrainNext is implemented in PyTorch and trained on a single NVIDIA A6000 GPU (48 GB memory) using mixed-precision (FP16) computation. Optimization is performed using the AdamW optimizer with a batch size of two owing to the computational requirements of three-dimensional volumetric learning. MRI volumes are processed using patch-based training, and He. initialization is employed for all convolutional layers. During inference, predictions from the five cross-validation models are ensembled to improve robustness and generalization across downstream tasks.

## 4. Experimental Setup

### 4.1 FOMO 2025 Benchmark

BrainNext was evaluated on the Foundation Models for Medical Imaging (FOMO) 2025 [28] Method Track, a large-scale benchmark designed to assess the generalization capability of self-supervised foundation models under limited-label and out-of-domain conditions. The Method Track provides a standardized evaluation protocol in which all participating methods are pretrained exclusively on the FOMO60K dataset containing 60,551 unlabeled brain MRI examinations, followed by adaptation to multiple downstream applications using limited labelled training data. This unified benchmark minimizes variability arising from different pretraining datasets and enables a fair comparison between foundation model architectures.

Unlike conventional evaluations that focus on a single downstream application, the FOMO benchmark assesses representation transfer across three distinct categories of medical image analysis: classification, segmentation, and regression. This design provides a comprehensive evaluation of learned representations while reflecting diverse clinically relevant neuroimaging tasks.

### 4.2 Downstream Tasks

The pretrained BrainNext encoder was independently fine-tuned for each downstream task using the labelled datasets provided by the FOMO challenge. A separate task-specific prediction head was optimized for each application while initializing the encoder with the pretrained MAE weights.

#### 4.2.1 Infarct Classification

The classification task aims to identify the presence of cerebral infarction from volumetric brain MRI examinations. The pretrained encoder is coupled with a fully connected classification head to produce binary predictions. Performance is evaluated using the Area Under the Receiver Operating Characteristic Curve (AUROC), which measures the model's ability to distinguish between infarct-positive and infarct-negative subjects across different decision thresholds.

#### 4.2.2 Meningioma Segmentation

The segmentation task focuses on voxel-wise delineation of intracranial meningiomas from three-dimensional MRI volumes. The pretrained BrainNext encoder is integrated with the U-Net decoder to generate dense segmentation masks. Performance is evaluated using the Dice Similarity Coefficient (DSC) and Normalized Surface Dice (NSD), which quantify volumetric overlap and boundary agreement between the predicted and reference segmentations.

#### 4.2.3 Brain Age Estimation

Brain-age estimation is formulated as a regression problem in which the model predicts chronological age directly from structural brain MRI. The pretrained encoder is combined with a regression head consisting of global average pooling and fully connected layers. Performance is assessed using the Mean Absolute Error (MAE) and the Pearson correlation coefficient (CORR) between predicted and chronological age.

A summary of the downstream tasks, learning paradigms, model outputs, and corresponding evaluation metrics is provided in Table 2.

### 4.3 Evaluation Metrics

Different evaluation metrics were employed according to the downstream task.

For infarct classification, model discrimination was assessed using the Area Under the Receiver Operating Characteristic Curve (AUROC). AUROC summarizes the trade-off between sensitivity and specificity across all classification thresholds and is widely used for binary medical image classification.

For meningioma segmentation, segmentation quality was quantified using the Dice Similarity Coefficient (DSC) [25] and the Normalized Surface Dice (NSD) [26]. DSC measures volumetric overlap between predicted and reference segmentations, whereas NSD evaluates agreement between object boundaries within a predefined tolerance, providing complementary assessments of segmentation accuracy.

For brain-age estimation, predictive accuracy was evaluated using the Mean Absolute Error (MAE) together with the Pearson correlation coefficient (CORR). MAE measures the average absolute difference between predicted and chronological age, while CORR assesses the linear relationship between predicted and true age across the study population.

### 4.4 Comparison with Competing Methods

BrainNext was evaluated under the official FOMO 2025 Method Track protocol and compared against all participating foundation models using the challenge evaluation server. All competing methods followed the same benchmark protocol, enabling direct comparison across the three downstream tasks without differences in pretraining data or evaluation methodology. Performance was assessed independently for infarct classification, meningioma segmentation, and brain-age estimation, together with the overall Method Track ranking.

Table 2. Summary of the downstream tasks and evaluation metrics used in the FOMO 2025 Method Track.

| Tasks | Learning Type | Output | Evaluation Metric |
|---|---|---|---|
| Infarct Classification | Classification | Binary label | AUROC |
| Meningioma Segmentation | Segmentation | Voxel-wise mask | DSC, NSD |
| Brain Age Estimation | Regression | Continuous age | MAE, CORR |

## 5. Experimental Results

### 5.1 Overall Performance

BrainNext was evaluated on the official Foundation Models for Medical Imaging (FOMO) 2025 Method Track, a standardized benchmark designed to assess the transferability of self-supervised foundation models across multiple neuroimaging applications under limited-label and out-of-domain conditions. The proposed framework was submitted under the team name Dolphins_creators and achieved second place overall in the Method Track. Unless otherwise stated, all results reported in this section correspond to the official blind test set evaluation provided by the FOMO 2025 challenge.

The quantitative performance of BrainNext across the three downstream tasks is summarized in Table 3, while the official FOMO 2025 leaderboard is presented in Figure 2.

Table 3. Performance of BrainNext on the official FOMO 2025 Method Track test set. Higher values indicate better performance for AUROC, DSC, NSD, and CORR, whereas lower values indicate better performance for MAE.

| **Downstream Task** | **Evaluation Metric** | **BrainNext** |
|---|---|---|
| Infarct Classification | AUROC ↑ | **0.684** |
| Meningioma Segmentation | DSC ↑ | **0.261** |
| | NSD ↑ | **0.232** |
| Brain Age Estimation | MAE ↓ | **12.67** |
| | CORR ↑ | **0.448** |
| **Overall Method Track Ranking** | Official Leaderboard | **2nd Place** |

BrainNext ranked first in meningioma segmentation and second overall, demonstrating that a single pretrained encoder can effectively transfer across classification, segmentation, and regression within a unified evaluation protocol.

**Track 1: Method**

| Placement | Team | Overall Rank | Task 1: Infarct Detection | | Task 2: Meningioma Segmentation | | | Task 3: Brain Age Estimation | | |
|---|---|---|---|---|---|---|---|---|---|---|
| | | | Rank | AUCROC | Rank | Dice | NSD | Rank | Correlation | MAE |
| **1** | **FOMO2JOMO** | **2.7** | **2** | **0.782** | **5** | **0.095** | **0.071** | **1** | **0.651** | **10.96** |
| **2** | **Dolphins_creators** | **4.8** | **7** | **0.684** | **1** | **0.261** | **0.232** | **6.5** | **0.448** | **12.67** |
| **3** | **TUKE** | **5.2** | **4** | **0.75** | **8** | **0.046** | **0.043** | **3.5** | **0.602** | **11.96** |
| **3** | **MaastrichtU-CDS** | **5.2** | **6** | **0.688** | **2.5** | **0.181** | **0.126** | **7** | **0.599** | **16.17** |
| 5 | FARM | 5.3 | 5 | 0.715 | 4 | 0.148 | 0.113 | 7 | 0.531 | 13.41 |
| 6 | DKFZ_MIC | 5.8 | 9 | 0.569 | 2.5 | 0.171 | 0.155 | 6 | 0.557 | 13.31 |
| 7 | MIPLAB | 6.0 | 10 | 0.473 | 6 | 0.065 | 0.053 | 2 | 0.632 | 11.48 |
| 8 | prenuvo | 6.2 | 8 | 0.649 | 7 | 0.064 | 0.047 | 3.5 | 0.626 | 12.53 |
| 9 | fomo25_brain | 6.8 | 3 | 0.76 | 9 | 0.006 | 0.005 | 8.5 | 0.409 | 14.85 |
| 10 | ashash | 7.2 | 1 | 0.789 | 10.5 | 0.003 | 0.004 | 10 | 0.37 | 17.62 |
| 11 | biopet | 10.8 | 11 | 0.44 | 10.5 | 0.005 | 0.002 | 11 | 0.137 | 18.21 |
| - | *baseline_unet_b* | - | - | *0.536* | - | *0* | *0* | - | *0.538* | *12.05* |
| - | *baseline_unet_xl* | - | - | *0.624* | - | *0* | *0* | - | *0.446* | *12.82* |

Figure 2. Official leaderboard of the FOMO 2025 Method Track highlighting the submission of BrainNext (team: Dolphins_creators), which achieved second place overall under the official challenge evaluation protocol [29].

As shown in Figure 2, BrainNext achieved the highest ranking for the meningioma segmentation task while maintaining competitive performance across infarct classification and brain-age estimation, resulting in an overall second-place finish in the Method Track.

### 5.2 Task-wise Performance

BrainNext demonstrated competitive performance across all three downstream tasks evaluated in the FOMO benchmark. For infarct classification, the pretrained foundation model achieved an AUROC of 0.684, indicating effective transfer of learned volumetric representations to binary disease classification.

For meningioma segmentation, BrainNext achieved a Dice Similarity Coefficient (DSC) of 0.261 together with a Normalized Surface Dice (NSD) of 0.232, demonstrating the ability of the pretrained encoder to support dense voxel-wise prediction through lightweight task-specific fine-tuning.

For brain-age estimation, BrainNext obtained a Mean Absolute Error (MAE) of 12.67 years and a Pearson correlation coefficient (CORR) of 0.448, illustrating that the learned anatomical representations generalize effectively to continuous regression tasks.

Overall, these results demonstrate that a single pretrained BrainNext encoder can be successfully adapted to three fundamentally different categories of medical image analysis while maintaining competitive performance under the standardized FOMO evaluation protocol.

## 6. Discussion

This study presented BrainNext, a general-purpose self-supervised foundation model [14, 15, 27] for volumetric brain MRI analysis based on masked autoencoder (MAE) pretraining and a native three-dimensional Bi-Directional xLSTM-UNet architecture. Unlike conventional supervised approaches that require independent models for individual applications, BrainNext learns transferable volumetric representations from large-scale unlabeled brain MRI data and subsequently adapts the pretrained encoder to multiple downstream tasks using lightweight task-specific prediction heads. Evaluation on the official FOMO 2025 Method Track demonstrated that a single pretrained encoder can effectively support classification, segmentation, and regression, achieving second place overall under the standardized challenge evaluation protocol. Although BrainNext was developed within the constraints of the FOMO benchmark, its design is independent of any specific downstream task and can readily be adapted to other neuroimaging applications through lightweight fine-tuning of the pretrained encoder.

The results highlight the effectiveness of large-scale self-supervised learning for volumetric neuroimaging. By leveraging more than 60,000 unlabeled brain MRI examinations during pretraining, BrainNext learns anatomically meaningful representations without relying on manual annotations. The competitive performance observed across three fundamentally different downstream tasks suggests that the learned features capture both local structural characteristics and global anatomical context, enabling effective transfer to diverse neuroimaging applications. These findings further support the growing role of foundation models as reusable feature extractors capable of reducing the dependence on large task-specific annotated datasets.

A distinguishing characteristic of BrainNext is the integration of a native three-dimensional Bi-Directional xLSTM-UNet architecture within the MAE framework. While conventional convolutional architectures primarily capture local spatial information, the bidirectional xLSTM modules facilitate modelling of long-range dependencies throughout the volumetric brain MRI. Combining hierarchical convolutional feature extraction with sequence modelling allows BrainNext to preserve fine anatomical details while simultaneously learning global contextual relationships, which is particularly important for complex neuroimaging tasks.

Despite these promising results, several limitations should be acknowledged. First, the current evaluation was performed exclusively within the FOMO 2025 benchmark. Although FOMO provides a standardized and comprehensive evaluation across multiple downstream tasks, additional validation on independent external datasets will be necessary to further assess the robustness and generalizability of BrainNext across different clinical populations, imaging protocols, and scanner vendors. Second, the downstream evaluation was limited to three representative neuroimaging tasks. Future work will investigate the transferability of BrainNext to additional applications, including Alzheimer's disease classification, brain tumour segmentation, lesion detection, image synthesis, and longitudinal disease progression modelling. Finally, although self-supervised pretraining substantially reduces annotation requirements, large-scale pretraining remains computationally demanding and requires significant GPU resources.

Future research will focus on scaling BrainNext using substantially larger neuroimaging datasets, exploring multimodal MRI pretraining, and investigating more advanced self-supervised objectives beyond masked image reconstruction. We also plan to evaluate BrainNext across broader neuroimaging benchmarks and release the pretrained model weights and inference code to facilitate reproducible research and encourage further development of foundation models for brain MRI analysis.

## 7. Conclusion

We presented BrainNext, a general-purpose self-supervised foundation model for volumetric MRI analysis. BrainNext combines masked autoencoder pretraining with a native three-dimensional Bi-Directional xLSTM-UNet architecture to learn transferable anatomical representations from 60,551 unlabeled brain MRI examinations without requiring manual annotations during pretraining. The pretrained encoder was subsequently adapted to infarct classification, meningioma segmentation, and brain-age estimation within the official FOMO 2025 Method Track.

Experimental results demonstrated that BrainNext effectively transfers learned volumetric representations across the three principal categories of medical image analysis: classification, segmentation, and regression, achieving second place overall under the official FOMO 2025 challenge evaluation protocol. These findings demonstrate the potential of self-supervised foundation models to provide a unified representation learning framework for diverse neuroimaging applications while reducing the need for separate task-specific models.

BrainNext provides a scalable foundation for future research in large-scale self-supervised learning for neuroimaging. Future work will investigate larger-scale pretraining, multimodal MRI foundation models, and broader clinical validation to further improve the generalizability of BrainNext. To support reproducible research and accelerate progress in the field, the pretrained BrainNext model weights and inference code will be made publicly available following publication.